\documentclass[10pt,twocolumn,letterpaper]{article}

\usepackage{iccv}
\usepackage{times}
\usepackage{epsfig}
\usepackage{graphicx}
\usepackage{amsmath}
\usepackage{amssymb}
\usepackage{float}
\usepackage{caption}
\usepackage{subcaption}

\usepackage[pagebackref=true,breaklinks=true,letterpaper=true,colorlinks,bookmarks=false]{hyperref}

\iccvfinalcopy 

\def\iccvPaperID{1129} 

\ificcvfinal\fi
\begin{document}

\title{Learning and Fusing Multimodal Features from and for Multi-task Facial Computing}

\author{Wei Li\\
CUNY City College\\
{\tt\small wli3@ccny.cuny.edu}
\and
Zhigang Zhu\\
CUNY City College and Graduate Center\\
{\tt\small zhu@cs.ccny.cuny.edu}
}

\maketitle

\begin{abstract}
We propose a deep learning-based feature fusion approach for facial computing including face recognition as well as gender, race and age detection. Instead of training a single classifier on face images to classify them based on the features of the person whose face appears in the image, we first train four different classifiers for classifying face images based on race, age, gender and identification (ID). Multi-task features are then extracted from the trained models and cross-task-feature training is conducted which shows the value of fusing multimodal features extracted from multi-tasks. We have found that features trained for one task can be used for other related tasks. More interestingly, the features trained for a task with more classes (e.g. ID) and then used in another task with fewer classes  (e.g. race) outperforms the features trained for the other task itself.  The final feature fusion is performed by combining the four types of features extracted from the images by the four classifiers. The feature fusion approach improves the classifications accuracy by a 7.2\%, 20.1\%, 22.2\%, 21.8\%  margin, respectively, for ID, age, race and gender recognition, over the results of single classifiers trained only on their individual features. The proposed method can be applied to applications in which different types of data or features can be extracted. 
\end{abstract}

\section{Introduction}

Facial computing, mainly face recognition, is the process of automatically identifying or verifying a person from an image or video frame and has been an active research area for a long time. There are mainly two types of approaches to do face recognition.  The first one is on the basis of designing image-specific features to represent the images and then applying common classifiers. The second group of methods, which have recently become very common, utilize deep learning methods. Deep learning models such as Convolutional Neural Networks (CNNs) ~\cite {krizhevsky2012imagenet} have shown high accuracy in face recognition. Although deep learning performs much better than most of the common machine learning and feature extraction approaches, researchers are still trying to improve these models by optimizing the training or fine-tuning these methods. 

A different approach to improving the performance of deep learning models is by using multiple channels or modalities when available in the training data. In most feature fusion methods, the feature fusion aims to convert a single channel to multiple channels either by data augmentation or network structure modification. For most multi-channel learning approaches, an extra source of features could provide additional information.  However, using only single-frame face images makes finding extra sources of information a big challenge.

Face recognition is the main task in facial computing. In addition to face recognition and verification, there have also been efforts for using the face images for facial expression, race or gender detection. These classification tasks work by training models and extracting features or descriptors that are suitable for particular tasks.  In this paper, we present a feature fusion approach to achieve higher accuracy through deep learning models. Instead of focusing on challenging the state of the art and improving existing face recognition algorithms, we try to prove the power of fused features automatically extracted by deep learning models for multiple tasks, and then used for each of the tasks.  The rationale is that it is possible that the feature extracted for one classification task may have useful information that another type of features extracted for a different classification task misses.  This leads to the assumption that we may have better classification results if we combine the features from different classifiers, even though the classifiers classify the data based on different criteria. Our approach is different with multi-task learning in the sense that we do not change the generation of original primary task feature.

To verify this hypothesis, we train multiple classification models using CNNs, all on the same data, but labeled differently. The classification criteria, which determine the labels, include person identification (ID), gender, race, and age recognition. Feeding an input image to the four models, we can extract four types of features, which can be fused to generate a more powerful feature for any of the four tasks (ID, gender, race and age recognition).

Since our goal is not proposing a new approach to outperform the state of the art work of face recognition, we are not focusing on designing and trying new structures for the deep learning models. The deep learning models we use in this work are CNNs. We perform multiple comparison experiments, which all show that the fusion features improve face recognition and all other recognition accuracy compared to the original features. 

The rest of the paper is organized as follows. Section 2 discusses related work. In Section 3 we explain our feature fusion approach and propose our hypothesis of multi-task features, after a brief introduction of the feature learning models we use. In Section 4 we describe the methods and experiments in multimodal feature learning for multiple tasks.  Then in Section 5, we present the feature fusion approach and the experiments.  We conclude and explain our future work in section 5.
\section{Related work}
In most applications where the data consists of images, the images are directly used to extract features. Some researchers focus on different patches of images to get enhanced representations.  In ~\cite {le2013multiple}, the authors propose to feed different patches of digital images into the model to extract the prediction vectors. By taking different regions in the original image into consideration, a fusion method was proposed to combine the region generated features. Then, a final prediction is acquired. This method divides the original image into several patches and then combines patch features based on patch region and weights to generate fused features. 

In order to do video classification, Machael, et al. ~\cite{guironnet2005video}  proposed several low-level descriptor models to extract high-level concepts. Low-level features such as color or texture are used to form fusion sensor to modify prediction based on Transferable Belief Model (TBM).  In addition to sensor fusion, the authors also propose concept (classification labels) fusion when relation exists between two concepts. 

A CNN based video classification study is introduced in ~\cite{karpathy2014large}, which makes use of the temporal information of the videos. Based on the fact that a video clip contains several contiguous frames in time, early fusion, late fusion and slow fusion of the video frames are proposed. The three fusion approaches are different in the structure of convolutional neural cells and the number of frames the model processes simultaneously. Their experiments show single frame with multi-resolutions and temporal fusion of multiple frames with complicated structures to achieve the best results.  The proposed approach improves the CNN model by modify the convolutional layer details based on the type of the input data and provides augmented input data (multi-resolution).  

Another important approach for fusion features and models are MTL (multi-task learning), which means applying one set of data to multi-task learning including the primary task the dataset designed for. Performance for the primary task will benefit from multi-task training. Richard Caruana ~\cite{caruna1993multitask} first introduced the idea of MTL, the training model are optimized by multiple label and corresponding loss, prediction results are supposed to be better when taking consideration of several kinds of loss. The MTL was applied into deep neural network for improving phoneme recognition in ~\cite{seltzer2013multi}. To enrich the feature for acoustic state prediction, the author added several other tasks of recognition like phone labels and state context. Multi-task learning DNN-HMMs are employed to combined the primary learning task and the secondary tasks respectively, and the best matching are selected among the several multi-task learning.  Collerbert and Weston ~\cite{collobert2008unified} proposed a deep network with multitask learning to the natural language processing task. A sentence are given several labels, experiments shows that multitasks learning always act better than single task and the more tasks it trains, the better performance it have.  MTLs are also applied to face verification to avoid overfitting when training on a small dataset ~\cite{wang2009boosted}. 

\section{Our Feature Learning and Fusion Approach}
\subsection{Learning through CNN}
Deep learning models have become very popular in image and signal-based classification. Compared to traditional classifiers, deep models have shown higher accuracy in recognition and detection ~\cite{sun2014deep, taigman2014deepface}. The accuracy of a deep learning approach relies heavily on the parameter updating methods and the training data. If the data presented to the model is large enough and also is a good representative of the entire data space, or if the training methods are more reliable, the accuracy can be improved. 

One way to improve the accuracy from the dataset side is to augment the data by applying transformations to it, which increases the size of the dataset. To improve the model itself, optimization has to be performed on the structure of the model and the training and fine-tuning algorithms. 

The deep learning model that we use in this work is the CNN. CNNs have received a lot of attention in the past few years, specifically in the tasks where the data is a set of images. Training a CNN includes two important parts: the forward pass estimation and the back-propagation mechanism. Contents of a CNN are mainly filter parameters and fully-connected layers. During the training part, the forward pass takes an incoming image as input and computes the classification prediction. The prediction is compared with the true class label and estimation error, also called the loss, is calculated. Then the back-propagation step uses the loss to update the parameters of the model. This process is repeated iteratively until the model converges and the loss is minimized. 

\begin{figure}[t]
\begin{center}
\includegraphics[width=0.8\linewidth]{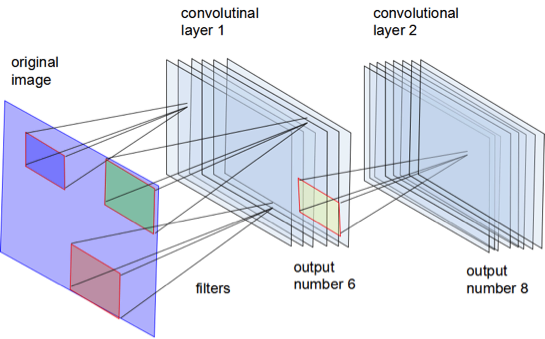}
\end{center}
\caption{The structure of a demonstrative convolutional neural network}
\label{fig1}
\end{figure}

The forward pass estimation part consists of multiple convolutional neural layers. There are various numbers of filters in each layer and output features are generated by passing input images through the filters. As shown in Figure \ref {fig1}, the first convolutional layer uses filters to operate on the original image. If the output is set to 6, there are 6 filters in first convolutional layer. The output of each layer will be the input to the next layer.
The back-propagation mechanism use the loss as input to fine-tune the whole CNN parameters. The convolutional filters are always initialized to Gaussian, Stochastic Gradient Decent (SGD) is employed to iteratively refine the weights of the layers. A learning rate is set during parameter update. Initially the value of the learning rate is set to a large number to increase the speed of convergence. Then the learning rate is gradually reduced to improve accuracy ~\cite{krizhevsky2012imagenet}. 

CNN can be used to extract features from the data ~\cite {jia2014caffe} . The features of an input image are in the forms of image or vectors that are the outputs of either the convolutional or the fully-connected layers , when feeding the input image into the CNN model. In low layers of the CNN (closer to the input layer), the features are more low-level and the retinotopical connection of the features to the original input images are notable. As we go deeper in the structure, the features become more high-level and more suitable for the classification task. We use the output of the connected layer from the CNN model as the learned features for feature fusion.  
\subsection{Overview of our  approach}
The representations learned by a deep learning model are similar to the ways human brain neurons learn representations of the sensory input. Our facial recognitions approach is also inspired by the way humans recognize faces. When we recognize or memorize people's faces, we are not only just recognizing the face and associating that to the identification, but there is always some other important information, such as the hairstyle, dressing style, facial expression, age, race, etc., which  can play an important role in identifying the person, even without us paying intentional attention to them. In our approach, we aim to use such additional information, including race, gender and age. 

\begin{figure}[t]
\begin{center}
\includegraphics[width=1\linewidth]{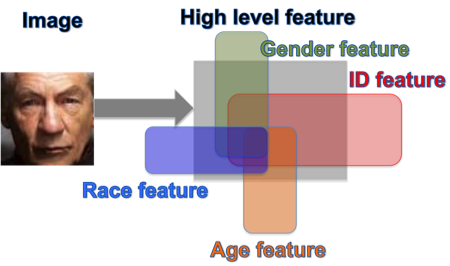}
\end{center}
\caption{Distribution of multi-task features in high-level space}
\label{fig2}
\end{figure}

The most effective way to obtain this information is to train separate CNN-based classifiers for race, gender and age models and extract and combine the features form these models. By intuition, the 3 types of features learned by different classifiers are suitable for different goals. Nevertheless we have found that features obtained from other individual classification tasks may have useful information that features extracted for a  certain classification task misses.  This leads to the proposed approach by which we may have better classification results if we combine the features from different classifiers, even though the classifiers classify the data based on different criteria. Figure \ref {fig2} shows the idea behind the proposed approach.

As shown in Figure \ref {fig2}, the image on the left is in RGB format and we can understand the image directly. In another space, we call it the high-level space, which the image information is in the format suitable for classifier to process. Assume the gray rectangle stands for the perfect high-level representation of the original image. While in deep learning training, we can get the certain task feature such as the ID feature or age feature if ID or age labels are available. Taking the ID feature as an example, the features we obtain by training a CNN classifier using the ID labels contain only parts of the  high level feature of the image. Otherwise the classification would be perfect. There are also some features out of the rectangle, which are for other specific tasks; these features are generated by comparing with others samples during training. Different features do not share information outside the high level rectangle since they do not know other task labels during training. Based on our hypothesis, we have the following hypotheses : 

1. Different features have overlapping areas, which means features for tasks of age or race classification can be used for identification, therefore we can run cross-task-feature recognition to verify it. 

2. The features trained for other task have the possibility to outperform its ''own'' feature obtained for that task, since the former feature may have a larger area in the common with the high-level feature. 

3. One feature may have an area inside the rectangle that other feature do not cover, therefore if multiple features are combined to do a classification task, the fusion feature may outperform individual features since more high level information is contained in the fusion feature. 

To verify whether our hypotheses are valid, we have run experiments in both multimodal feature learning and fusion, described in Sections 4 and 5, respectively.	
\section{CNN-based Multimodal Feature Learning}
As we described in Section 3, our goal is to combine all features learned from ID, gender, race and age detection. For this purpose, we need to train 4 classifiers with  ID, gender, age, and race labels. Since it would be too much work if we manually label an existing dataset, and even we do this, the numbers of samples may not be balanced (which will affect the training), we have collected our own dataset.

We group people by gender (male, female), age (young, adult, old) and race (Asian, Latin-American, African, White). Therefore we have 2x3x4=24 sub-groups. For each of the sub-group we identify several names of famous people, for whom it is easy to obtain a large number of images from the Web. A face detector ~\cite{pedregosa2011scikit} is used to extract the face area from each image, and then we manually went through the dataset to remove the wrong images. Finally we collected 24000 images for 77 people, each has over 300 images on average.

\begin{figure}[t]
\begin{center}
\includegraphics[width=1\linewidth]{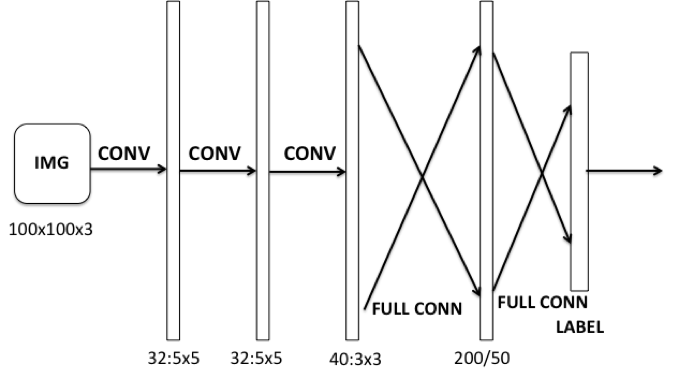}
\end{center}
\caption{Feature learning CNN structure}
\label{fig3}
\end{figure}

Since the ID, gender, age and race labels of the face image datasets are now available, we first train a CNNs classifier for each of the 4 tasks. The same CNN structure is used for all four classifiers, as shown in Figure \ref {fig3}. Each CNN network is composed of an input layer, 3 convolutional layers and 2 fully-connected transformation layers (a high-level feature extraction layer and an output layer for labels). The convolutional filters we use for the 3 layers are 5x5, 5x5 and 3x3, respectively.  The only difference of the  structures for the four tasks is that, in the CNN structure for ID, the number of features for the high-level feature extraction layer is defined as 200 since there are much more classes (77 versus 2,3 and 4), while this parameter is set to 50 in other CNNs. 
The number of images in the original dataset (24000) is not sufficient for training our CNN models. Therefore we augment the dataset 10 times by applying image filters and affine transforms. For the following experiments, we use 70\% of the augmented images for training and the remaining 30\% for testing. Based on the CNNs structures, we used 200 images for training in each iteration, and the total number of iterations is 20,000.

\begin{figure}[t]
\begin{center}
\includegraphics[width=1\linewidth]{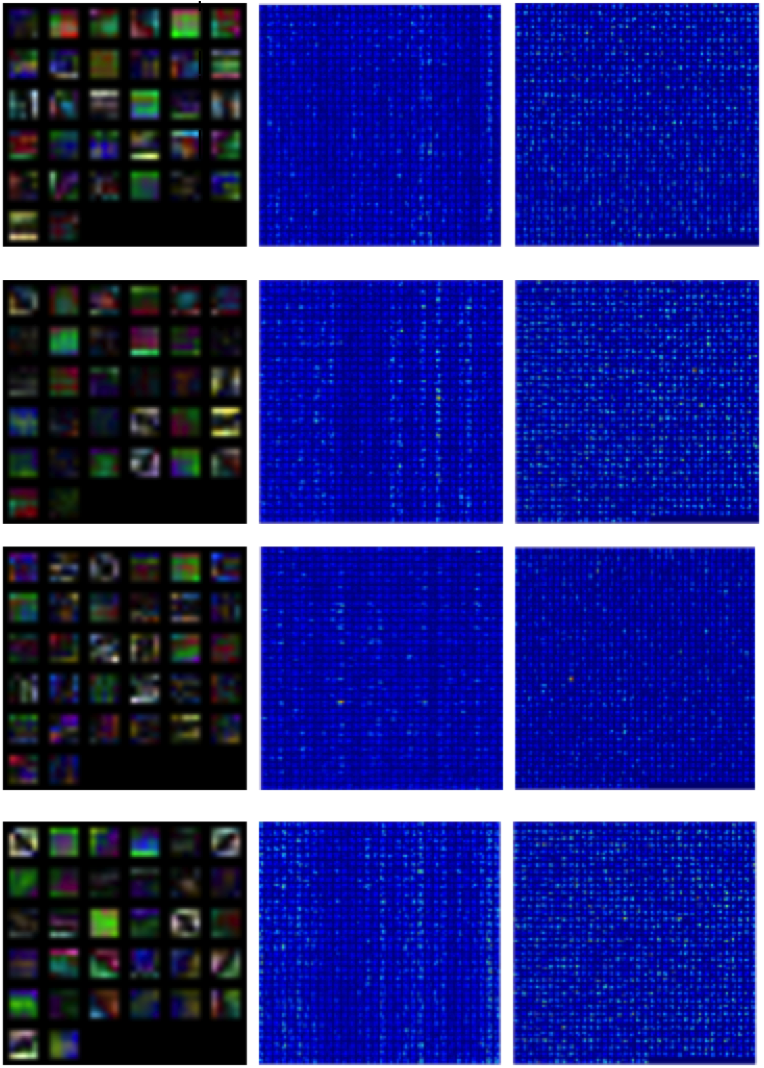}
\end{center}
\caption{Visualization of the trained layers (left to right are the 1st, 2nd 3rd convolutional layers, from top to bottom are the layers for ID, age, race and gender)}
\label{fig4}
\end{figure}

\begin{figure}[t]
\begin{center}
\includegraphics[width=1\linewidth]{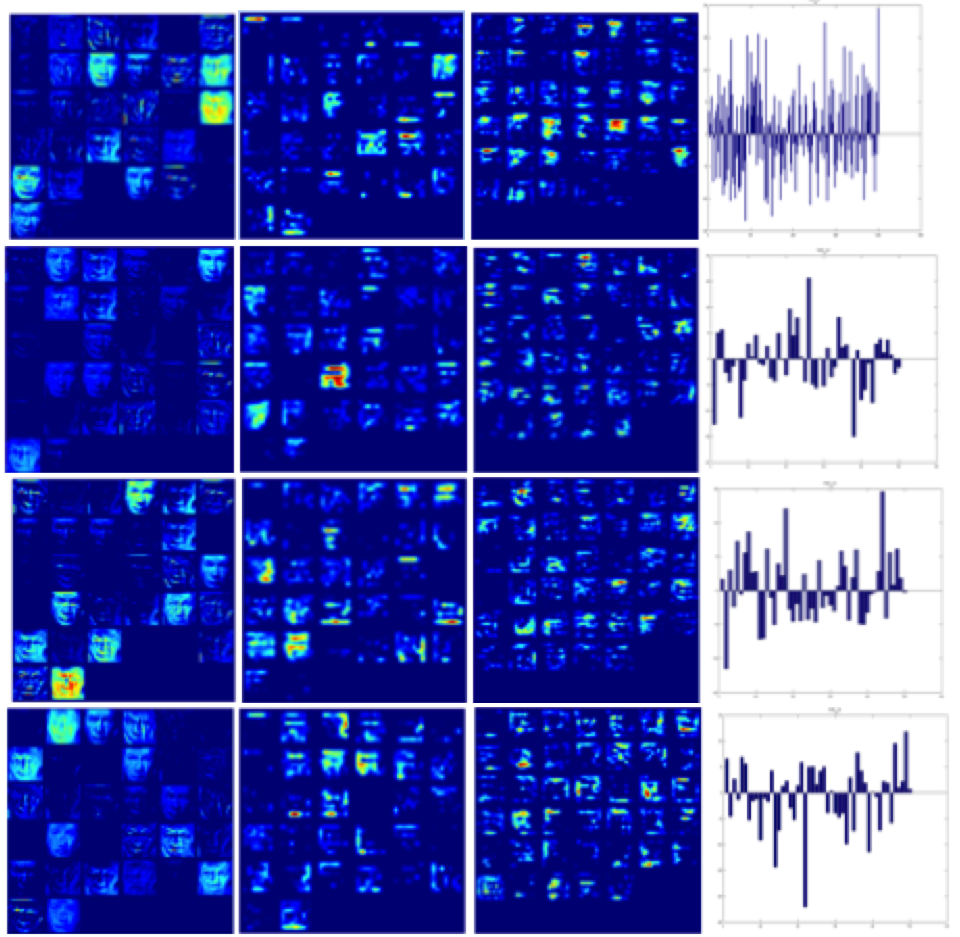}
\end{center}
\caption{Visualization of the features extracted from different layers of each task. (Rows 1 to 4: ID, age, race, gender)}
\label{fig5}
\end{figure}

The CNN structure enables us to extract features from the data in each layer. In Figure \ref {fig4}, we visualize the parameters of the trained models to show the differences of the four models. We also show the features of one of the sample images generated by the four models in Figure \ref {fig5}. The convolutional layers contain regular image processing filters, but by combining large number of these filters through the model, powerful predictions can be made by the CNNs.
Shown in Figure \ref {fig5} (from row 1 to row 4) are the features extracted from the three convolution layers (columns 1 to 3) and the high-level feature extraction layer (column 4) of the CNN models for ID, age, race and gender, respectively. From these features  we can still see that the three convolutional layers are mainly low-level feature extractors, while the high-level feature extraction layer indeed extracted higher level features of the image.  Compared to the low level features, the high-level feature is more descriptive and compact to use in cross-task classification below. 

Before we move to the feature fusion step, we want to see if features from one task can be useful in other tasks, as hypotheses 1 and 2 suggest. If this works, it  means the trained features have both common features about the images and the specific features for the classification tasks.

\begin{figure}[t]
\begin{center}
\includegraphics[width=1\linewidth]{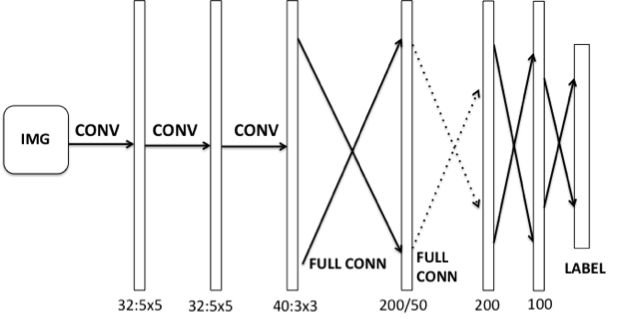}
\end{center}
\caption{CNN structure for cross-task-feature recognitions}
\label{fig6}
\end{figure}

To test the features extracted for one task (e.g. ID) on other tasks (age, race or gender), we connect the high-level feature extraction layer of the trained CNN models for this task to two more fully-connected layers, before the final output layer,  as shown in Figure \ref {fig6}.  For each of the other tasks, e.g., for age classification, the two layers in the corresponding CNN are trained by feeding each image of the training set to the trained CNN model (for ID) to obtain the cross-task high-level features, and then using them as the input to the 2-layered full connected model.  Then the newly trained cross-task model (ID for age) is tested with the test set for each task (age in this example).

We apply individual features to all four tasks to perform the ''cross-task-feature recognition''. Table 1 shows the results of this experiment. We highlighted the 2 best performances for each task (if they are above 50\%). We can see that all classifiers performs well using their ''own'' features and for all the cross-feature recognition cases, the results are always much better than random guesses.  In particular, ID features do a much better job for other three ''simpler'' tasks like  age, race and gender recognition. 
\begin{table}
\caption{The cross-task-feature recognition results.}
\begin{center}
\begin{tabular}{lllll}
\hline
Features & ID(77) & Age(3) & Race(4) & Gender(2) \\
\hline\hline
ID(200) &\bf 60.2\% & \bf83.2\% & \bf83.0\% & \bf92.3\% \\
Age(50) & 23.6\% &\bf69.1\% & 58.4\% &76.9\% \\
Race(50) & 24.8\% & 62.3\% &\bf67.7\% &76.7\% \\
Gender(50) & 22.8\% & 63.1\% & 58.3\%& \bf84.3\%\\
\hline
\end{tabular}
\end{center}
\end{table}

Table 1 confirms our hypothesis about cross-feature recognition. Figure 8 gives more details of the changes of accuracy for the test data during the training of the cross-feature models. The accuracy always starts from random guessing then gradually reaches the final result. For the ID task, using features of the other three tasks can only achieve a 20\% to 25\% accuracy. Considering we have 77 classes to deal with and the features for the other three tasks are only trained to handle 2 to 4 classes, the results are still acceptable. For all the other three tasks, the gender, age and race recognition, their own features always have a sound prediction. More interestingly, the performance of the cross-feature recognition results are very close to the performance using individual features. Among them, it is very interesting to see that the features trained for ID have better performance in the age, race, and gender recognition than their own trained features. According to our feature distribution graph, we claim that the identification features have more high-level information to represent the original image. This may tell that more classifying categories will train the features deeper. So our insights is that when we have a small number of groups to classify, we can try to divide the groups into smaller ones, since in this way the loss function will help the network learn better features.  

\begin{table}
\caption{Fusion feature recognition result versus own and other 3 fusion feature}
\begin{center}
\begin{tabular}{lllll}
\hline
Features & ID & Age & Race & Gender \\
\hline\hline
Own & 60.2\% & 69.1\% & 67.7\% & 84.3\% \\
Other three & 45.3\% &87.3\% & 86.9\% &94.9\% \\
All & 67.4\% & 89.2\% &89.9\% &96.1\% \\
\hline
\end{tabular}
\end{center}

\end{table}

\begin{figure}[t]
\begin{center}
\includegraphics[width=1\linewidth]{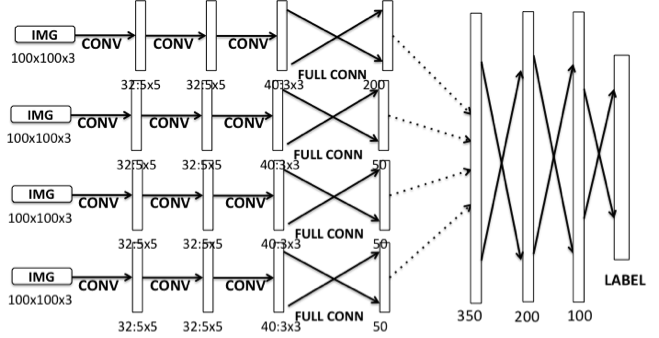}
\end{center}
\caption{CNN structure for the fused feature multi-tasks recognition}
\label{fig7}
\end{figure}


\section{Feature Fusion Approach and Experiments}
A good feature representation should be able represent most of the common features of representing the image as well as features specific to each individual classification tasks. Our cross-task-feature recognition experiments show that some of the common features can be learned by different tasks. It is very likely if we combine multiple features from multi-tasks to generate a fused feature for each of the tasks, we may have better classification for any of the classification tasks. For example, we can combine the ID feature with other three features (for age, race and gender) by concatenating them to a new one dimensional feature. As we have mentioned, the feature length for the ID task is 200 while the length for all the other 3 features is 50. Therefore we have a fused feature with a length of 350. We have also used fused features for all the other three tasks.  The CNN structure for this feature fusion is shown in Figure \ref {fig7}.  In the new addition, we have four layers: the first layer is simply the concatenating layer, and after two fully-connected layers, we have the output layer for labels. Again we trained these layers for each of the ID, age, race and gender tasks, using the fused features of the training dataset, and tested the fusion model for each task. The results is shown in Table 2.
\begin{figure}[t]
\begin{center}
\includegraphics[width=0.9\linewidth]{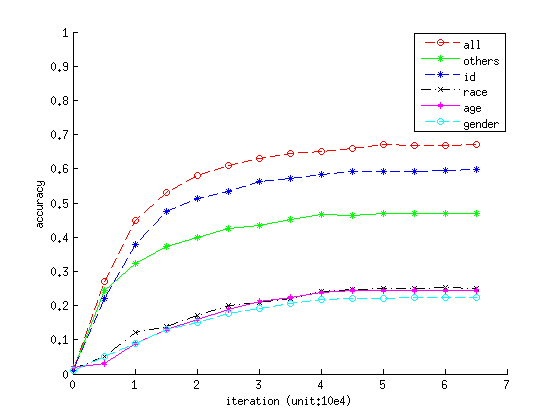}
\end{center}
\caption{Comparison testing results on ID task training}
\label{idfig}
\end{figure}

\begin{figure}[t]
\begin{center}
\includegraphics[width=0.9\linewidth]{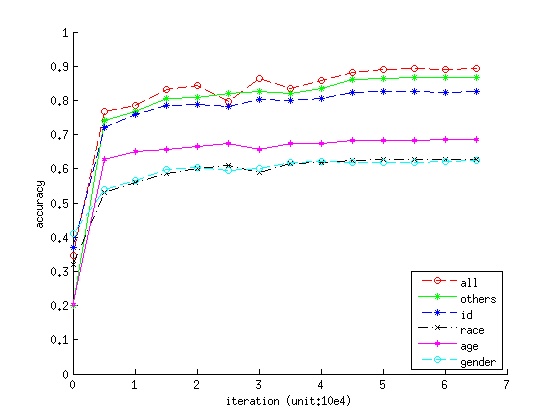}
\end{center}
\caption{Comparison testing results on Age task training}
\label{agfig}
\end{figure}

\begin{figure}[t]
\begin{center}
\includegraphics[width=0.9\linewidth]{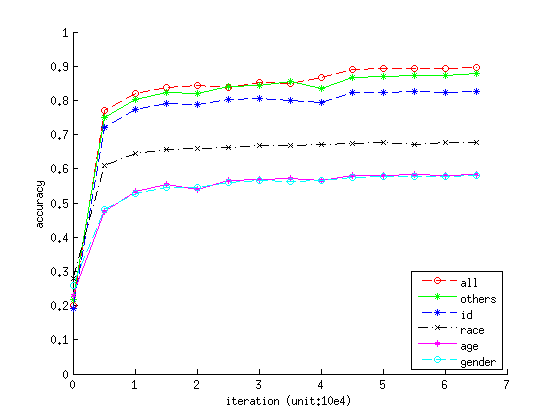}
\end{center}
\caption{Comparison testing results on Race task training}
\label{etfig}
\end{figure}

\begin{figure}[t]
\begin{center}
\includegraphics[width=0.9\linewidth]{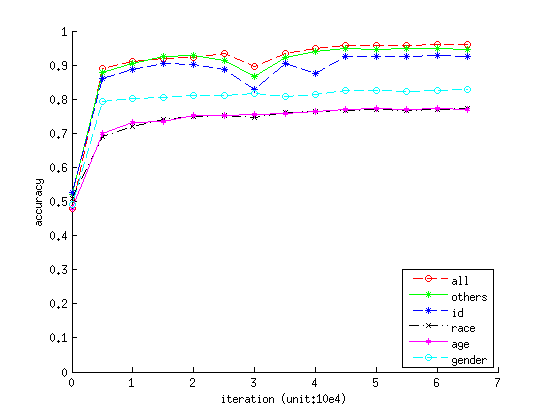}
\end{center}
\caption{Comparison testing results on Gender task training}
\label{gdfig}
\end{figure}
From the table, we can see that the final result have 7.2\%, 20.1\%, 22.2\%, 21.8\%  percentage improvement to the original features with the same setting.  The combination of all the other three task features can also have big improvements in race, gender and age prediction, and a acceptable prediction for identification prediction. This indicates that we can even train features for a task without having training labels of that task. We have also recorded all the testing scores with testing dataset during the training iterations (Figure \ref {idfig} to Figure \ref{gdfig} ), which will have a more obvious description of the difference of the features we used. From Figure \ref {idfig} to Figure \ref{gdfig}, we can see the best performance is always by the fusion feature (red). The performance of their "own" feature (blue) ranks the 2nd in ID and does well in race, age and gender recognition. In the cross-task feature recognition of race, gender and age, the own feature act a little better than the other two, but the two other features are in the same level of performance in cross-task-feature recognition.

\section{Conclusion and Future Work}
In this paper, we proposed a feature learning and fusion approach for deep learning based facial computing. In addition to train a CNN model of training images for the ID recognition task, we trained three other  models for race, age and gender recognition. We did the cross-task-feature recognition across multiple tasks and exhibited the value of multiple task features. By combined the features extracted from other three models with the original feature, we obtain the fusion features, which lead to a 7.2\%, 20.1\%, 22.2\%, 21.8\% improvement in the ID , age, race and gender prediction, respectively.
 
In the future, we would like to leverage the finding that training a system using more sub-groups can boost the performance of the original task with fewer groups. This is even true for related but different tasks.  We plan to continue to work on to enrich our feature fusion approach. Right now, we only have sub-groups of race, gender and age; for some other groups  such as facial expression, the situation may be more complicated. However, we will see the possibility of applying feature fusion approach to this type of task. So far we are generating the fusion feature directly by concatenating singe feature, we would like to try different merging algorithms to see if improvement can be made. We will also see if our approach can contribute to the improvement to the state of the art in face recognition/verification.

{\small
\bibliographystyle{ieee}
\bibliography{fusion}

\begin{thebibliography}{10}\itemsep=-1pt

\bibitem{caruna1993multitask}
R.~Caruna.
\newblock Multitask learning: A knowledge-based source of inductive bias.
\newblock In {\em Machine Learning: Proceedings of the Tenth International
  Conference}, pages 41--48, 1993.

\bibitem{collobert2008unified}
R.~Collobert and J.~Weston.
\newblock A unified architecture for natural language processing: Deep neural
  networks with multitask learning.
\newblock In {\em Proceedings of the 25th international conference on Machine
  learning}, pages 160--167. ACM, 2008.

\bibitem{guironnet2005video}
M.~Guironnet, D.~Pellerin, and M.~Rombaut.
\newblock Video classification based on low-level feature fusion model.
\newblock In {\em European Signal Processing Conference (EUSIPCO'2005)},
  page~CD, 2005.

\bibitem{jia2014caffe}
Y.~Jia, E.~Shelhamer, J.~Donahue, S.~Karayev, J.~Long, R.~Girshick,
  S.~Guadarrama, and T.~Darrell.
\newblock Caffe: Convolutional architecture for fast feature embedding.
\newblock In {\em Proceedings of the ACM International Conference on
  Multimedia}, pages 675--678. ACM, 2014.

\bibitem{karpathy2014large}
A.~Karpathy, G.~Toderici, S.~Shetty, T.~Leung, R.~Sukthankar, and L.~Fei-Fei.
\newblock Large-scale video classification with convolutional neural networks.
\newblock In {\em Computer Vision and Pattern Recognition (CVPR), 2014 IEEE
  Conference on}, pages 1725--1732. IEEE, 2014.

\bibitem{krizhevsky2012imagenet}
A.~Krizhevsky, I.~Sutskever, and G.~E. Hinton.
\newblock Imagenet classification with deep convolutional neural networks.
\newblock In {\em Advances in neural information processing systems}, pages
  1097--1105, 2012.

\bibitem{le2013multiple}
H.~M. Le, A.~T. Duong, and S.~T. Tran.
\newblock Multiple-classifier fusion using spatial features for partially
  occluded handwritten digit recognition.
\newblock In {\em Image Analysis and Recognition}, pages 124--132. Springer,
  2013.

\bibitem{pedregosa2011scikit}
F.~Pedregosa, G.~Varoquaux, A.~Gramfort, V.~Michel, B.~Thirion, O.~Grisel,
  M.~Blondel, P.~Prettenhofer, R.~Weiss, V.~Dubourg, et~al.
\newblock Scikit-learn: Machine learning in python.
\newblock {\em The Journal of Machine Learning Research}, 12:2825--2830, 2011.

\bibitem{seltzer2013multi}
M.~L. Seltzer and J.~Droppo.
\newblock Multi-task learning in deep neural networks for improved phoneme
  recognition.
\newblock In {\em Acoustics, Speech and Signal Processing (ICASSP), 2013 IEEE
  International Conference on}, pages 6965--6969. IEEE, 2013.

\bibitem{sun2014deep}
Y.~Sun, X.~Wang, and X.~Tang.
\newblock Deep learning face representation from predicting 10,000 classes.
\newblock In {\em Computer Vision and Pattern Recognition (CVPR), 2014 IEEE
  Conference on}, pages 1891--1898. IEEE, 2014.

\bibitem{taigman2014deepface}
Y.~Taigman, M.~Yang, M.~Ranzato, and L.~Wolf.
\newblock Deepface: Closing the gap to human-level performance in face
  verification.
\newblock In {\em Computer Vision and Pattern Recognition (CVPR), 2014 IEEE
  Conference on}, pages 1701--1708. IEEE, 2014.

\bibitem{wang2009boosted}
X.~Wang, C.~Zhang, and Z.~Zhang.
\newblock Boosted multi-task learning for face verification with applications
  to web image and video search.
\newblock In {\em computer vision and pattern recognition, 2009. CVPR 2009.
  IEEE conference on}, pages 142--149. IEEE, 2009.

\end{thebibliography}
}

\end{document}